\documentclass[runningheads]{llncs}
\usepackage{graphicx}
\usepackage{xcolor}
\usepackage{graphicx} 
\usepackage{booktabs}
\usepackage{multirow}
\usepackage{adjustbox}
\usepackage{subfigure}
\usepackage{amsmath}
\usepackage{hyperref}

\newcommand{\eat}[1]{}

\renewcommand{\thefootnote}{[\arabic{footnote}]}
\usepackage{multirow}
\begin{document}

\title{InvisibiliTee: Angle-agnostic Cloaking from Person-Tracking Systems with a Tee}
%\title{InvisibiliTee: Angle-agnostic Cloaking from Person-Tracking Systems with a Tee\thanks{Supported by organization x.}}
%
\titlerunning{InvisibiliTee}

\author{Yaxian Li\inst{1,2} \and
Bingqing Zhang\inst{1,2} \and
Guoping Zhao\inst{1,2} \and
Mingyu Zhang\inst{1,2} \and
Jiajun Liu\inst{4} $^{*}$ \and
Ziwei Wang\inst{4} \and
Jirong Wen \inst{1,2,3}
}
\renewcommand{\thefootnote}{\fnsymbol{footnote}}
\footnotetext[1]{Corresponding author. Jiajun.liu@csiro.au}

%\author{First Author\inst{1}\orcidID{0000-1111-2222-3333} \and
%Second Author\inst{2,3}\orcidID{1111-2222-3333-4444} \and
%Third Author\inst{3}\orcidID{2222--3333-4444-5555}}
%
\authorrunning{Yaxian Li, Bingqing Zhang. et al.}
\institute{School of Information, Renmin University of China, Beijing, China \and
Beijing Key Laboratory of Big Data Management and Analysis Methods, Beijing, China \and
Gaoling School of Artificial Intelligence, Renmin University of China, Beijing, China \and
Data 61, CSIRO, Pullenvale, Australia
}
\maketitle              % typeset the header of the contribution
\begin{abstract}
After a survey for person-tracking system-induced privacy concerns, we propose a black-box adversarial attack method on state-of-the-art human detection models called \textbf{InvisibiliTee}. The method learns printable adversarial patterns for T-shirts that cloak wearers in the physical world in front of person-tracking systems. We design an angle-agnostic learning scheme which utilizes segmentation of the fashion dataset and a geometric warping process so the adversarial patterns generated are effective in fooling person detectors from all camera angles and for unseen black-box detection models. Empirical results in both digital and physical environments show that with the \textbf{InvisibiliTee} on, person-tracking systems' ability to detect the wearer drops significantly.\footnote{Code is available at https://github.com/invisibilitee/invisibilitee}

\end{abstract}

\vspace{-0.6cm}
\section{Introduction}
\label{sec:intro}
\vspace{-0.2cm}

Person-tracking systems are widely deployed in metropolitan areas across the world for various purposes. According to the latest Comparitech report \footnote{https://www.comparitech.com/vpn-privacy/the-worlds-most-surveilled-cities/}, approximately 770 million cameras have already been used globally, which include many smart cameras enabled with person-tracking systems. 
While making contributions to public safety, these network-enabled systems suffer from software/hardware vulnerabilities and are often prone to cyber-attacks 
%\footnote{https://www.bloomberg.com/news/articles/2021-03-09/hackers-expose-tesla-jails-in-breach-of-150-000-security-cams}
\footnote{https://www.scmp.com/news/china/article/1727145/chinese-surveillance-camera-supplier-confirms-hacking-loophole}, raising serious concerns about privacy breaches.
Thus, average citizens are now facing even stronger privacy risks. For instance, after hacking into camera-based person-tracking systems, it is not only possible but even unchallenging to precisely recover a person's daily routine. 

%\vspace{-0.2cm}
\begin{figure}[htbp]
    \centering
    \includegraphics[width=0.8\textwidth]{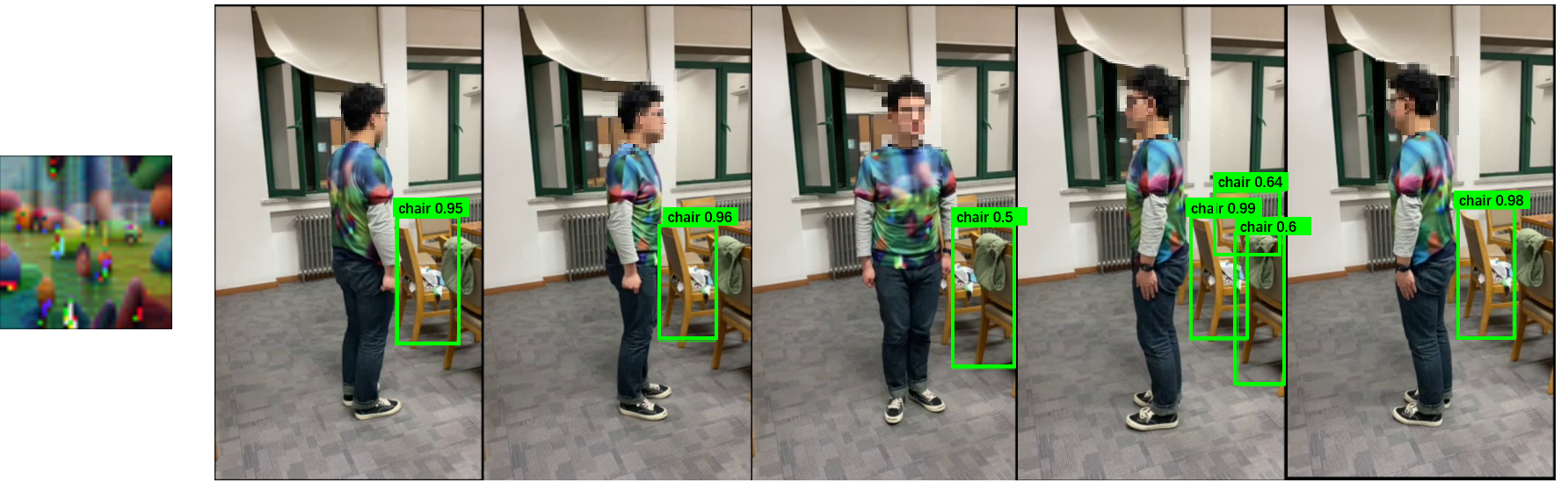} 
    \vspace{-0.3cm}
    \caption{An adversarial pattern and its physical attack results with a person wearing corresponding ``InvisibiliTee''. }
    \label{fig:head_eg}
    %\vspace{-0.5cm}
\end{figure}
% 
%\vspace{-0.55cm}
To learn how this issue is perceived by the general public, we conducted a small-scale questionnaire-based survey. Among the 20 participants, all are concerned with the person-tracking systems' potential breach of privacy and 90\% responded that they feel the need to leverage latest technology to protect personal privacy, as a counter-balance to the evolving person-tracking technology. 
%The pedestrians in public spaces should have the basic rights to decide whether or not they want to be identified and tracked by such systems. 
This interesting study inspires our research for the ``InvisibiliTee'', a tee that cloaks wearer in front of the tracking system.
As illustrated in Fig.~\ref{fig:head_eg}, people can successfully fool the person detection system with the ``InvisibiliTee'' on. Before diving into the technical details of the ``InvisibiliTee'', we first show how the survey is conducted and what we could conclude from the results.

\vspace{-0.4cm}
\subsection{User Study}
%\vspace{-0.6cm}

\vspace{-0.6cm}
\begin{figure}[htbp]
    \centering
    \includegraphics[width=0.6\textwidth]{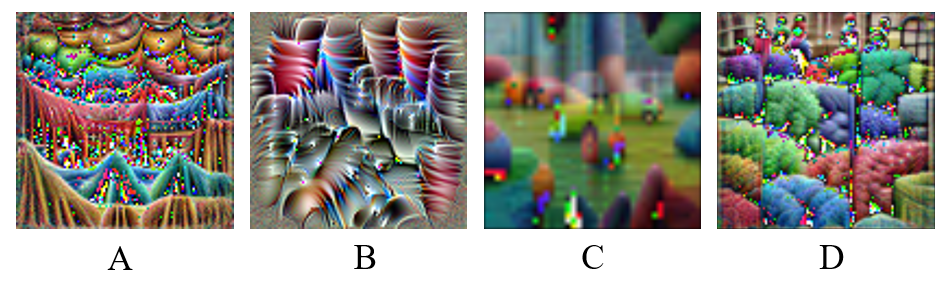}
    \vspace{-0.3cm}
    \caption{Adversarial patterns in the questionnaire. }
    \label{fig:question}
\end{figure}
\vspace{-0.4cm}

\begin{table}[htbp]
\small
\centering
\caption{Questions in the user study questionnaire.}
\resizebox{\textwidth}{!}
{
\begin{tabular}{c|l|l}
\toprule[1pt]
\textbf{ID} & \multicolumn{1}{c|}{\textbf{Question}}                                                                                                                                                                                                     & \multicolumn{1}{c}{\textbf{Choices}}                                                                                         \\ \hline
1               & \begin{tabular}[c]{@{}l@{}}Are you worried that cameras in public places\\ will infringe on personal privacy?\end{tabular}                                                                                                                & \begin{tabular}[c]{@{}l@{}}A. Don’t mind at all\\ B. Somewhat concerned\\ C. Concerned\\ D. Extremely concerned\end{tabular} \\ \hline
2               & \begin{tabular}[c]{@{}l@{}}Do you think there is a need to use technical \\means to protect personal privacy?\end{tabular}                                                                                                                & \begin{tabular}[c]{@{}l@{}}A.Yes \\ B. No\end{tabular}                                                                       \\ \hline
3               & \begin{tabular}[c]{@{}l@{}}If there is an ``InvisibiliTee'' to protect personal \\privacy, are you willing to wear it?\end{tabular}                                                               & \begin{tabular}[c]{@{}l@{}}A. Yes \\ B. No\\ C. Depends\end{tabular}                                                          \\ \hline
4               & \begin{tabular}[c]{@{}l@{}}The following pictures in Fig~\ref{fig:question} are different ``InvisibiliTee''  \\patterns. Please select the patterns that are \\acceptable. (multiple choices)\end{tabular} & \begin{tabular}[c]{@{}l@{}}A. Pattern A B. Pattern B\\ C. Pattern C D. Pattern D\\ E. Unacceptable\end{tabular}          \\
\bottomrule[1pt]
\end{tabular}
}
\label{tab:question}
\end{table}

\vspace{-0.2cm}
As shown in Table~\ref{tab:question}, there are 4 questions in the questionnaire. It involves the urgency of privacy issues, the acceptance of technical means and adversarial patterns. The 20 participants surveyed are college students aged between 22 to 30, with higher education background and living experience in metropolis. They are also the target demographic group that InvisibiliTee tries to help - younger generations with stronger awareness of privacy risks and open mindset towards early technology adoption.
Among the 20 people surveyed, all are worried about the camera's invasion of personal privacy, of which 55\% are concerned and 35\% are extremely concerned. 90\% say they feel the need to leverage technical approaches to protect personal privacy and over a half show the willingness to wear an `` InvisibiliTee''. 
There are 7 people who think the patterns of  `` InvisibiliTee'' not fashionable enough to wear. But others do consider these patterns acceptable, among which pattern C in Fig~\ref{fig:question} is most popular, chosen by 10 people. 
%Further details of the investigation can be found in supplementary material.

% this part finally comes into introducing the model.
Motivated by a strong need for privacy protection in the current world where person-tracking systems are often abused, an angle-agnostic black-box adversarial attack method, namely \textbf{InvisibiliTee}, is proposed in this paper. Its advantages are summarized as follows:
1, \textbf{Cross-model genenralizability.} The learned attack patterns achieve competitive black-box attack results on unseen detection models. 2,\textbf{Cross-scene/subject genenralizability.} The same learned adversarial pattern can be used by applied on unseen wearers in unseen scenes without re-training. 3, \textbf{Angle-agnostic.} In the digital world, adversarial patterns can be applied on images taken from different angles. In the physical world, we use fabric with fully printed adversarial pattern to tailor a T-shirt, achieving angle-agnostic ``invisible'' effect and qualitative analysis shows effectiveness of attack to some extent.

\eat{
\vspace{-0.3cm}
\begin{enumerate}
    \item \textbf{Cross-model genenralizability.} The learned attack patterns achieve competitive black-box attack results on unseen detection models.
    \vspace{-0.3cm}
    \item \textbf{Cross-scene/subject genenralizability.} The same learned adversarial pattern can be used by applied on unseen wearers in unseen scenes without re-training.
    \vspace{-0.3cm}
    \item \textbf{Angle-agnostic.} In the digital world, adversarial patterns can be applied on images taken from different angles. In the physical world, we use fabric with fully printed adversarial pattern to tailor a T-shirt, achieving angle-agnostic ``invisible'' effect.
\end{enumerate}
}

\vspace{-0.4cm}
\section{Literature Review} \label{sec:literature}
\vspace{-0.2cm}

In person tracking system, a critical step is to utilize object detection algorithms to generate bounding box to crop the images with persons.
Fast R-CNN~\cite{girshick2015fast} introduced a RoIPooling operation to accelerate detection speed while maintaining detection performance. He et al. proposed Faster R-CNN~\cite{ren2016faster}, utilizing Regional Proposal Network to generate proposals instead of time-consuming selective search~\cite{uijlings2013selective}. R-FCN~\cite{dai2016r} comes up with position-sensitive RoIPooling for further improving  efficiency.
Single-stage frameworks, such as YOLO~\cite{redmon2016you,redmon2017yolo9000,farhadi2018yolov3} and SSD~\cite{liu2016ssd,fu2017dssd,lin2017focal}, directly classify and regress bounding boxes. 

% attacks on detection models
Although detectors are getting better performances, recent studies have proven that the deep object detection models are vulnerable to attacks from adversarial examples~\cite{szegedy2014intriguing,goodfellow2014explaining}.
Adversarial attack methods can be divided into two categories, which are \textbf{white-box} attack and \textbf{black-box} attack.
In white-box attack, the attacker can interact with the machine learning system in the process of generating adversarial attack data, obtaining the gradient during the training process.
While in black-box attack, attackers do not know the algorithms nor parameters of the target model. But attackers can still observe the output through any input.
Several attempts have been made on attacking Re-ID systems via black-box methods~\cite{Wang2020CVPR,Zheng2020Reid}. 
Although existing works demonstrate the feasibility of performing black-box attacks on classification and Re-ID vision systems, the research on person detection system has not yet been sufficiently investigated.
\eat{However, these assumptions of white-box attacks are impractical in real life.
Firstly, the person detection model directly obtains the image from the camera instead of the fraudulent image elaborately designed by attackers. Secondly, it is often the case that attackers have no prior knowledge of the target model. 
On the contrary, black-box attacks~\cite{su2019one} that requires minimal knowledge about the target detection models demonstrate more realistic properties.}

\eat{
% Challenges of blackbox attack
The nature of black-box attacks on person detectors poses further challenges. 
Firstly, the amount of available information is far less in black-box attacks than in white-box counterparts.
Secondly, in a natural open scene, the lighting conditions, the angle of the object, and the occlusion bring great challenges to adversarial attacks on the images collected by the camera.}

%Recently, adversarial attack on object detection has aroused wide attention.
Liu et al.~\cite{liu2019dpatch} propose DPATCH, which is a universal black-box adversarial attack method. It borrows the idea of adversarial patch~\cite{brown2017adversarial} that simultaneously attacks bounding boxes regression and classification. 
Later, Lee et al.~\cite{lee2019on} consider a normalized steepest ascent approach, PGD~\cite{madry2018towards}, to update loss function. Thys et al.~\cite{thys2019fooling} try to attack the object with high level of intra-class variety such as person. 
Recently, Wu et al.~\cite{wu2020making} design an ``invisibility cloak'' that decreases the scores of a series of object detector. 
However, to the best of our knowledge, only a few works~\cite{wu2020making} focused on designing a pattern on a T-shirt which can attack detectors without direction restrictions. We propose ``InvisibiliTee'', which is an angle-agnostic adversarial attack method.

\vspace{-0.5cm}
\section{Method} \label{sec:method}
\vspace{-0.7cm}

\begin{figure*}[ht]
    \centering
    \includegraphics[width=0.9\textwidth]{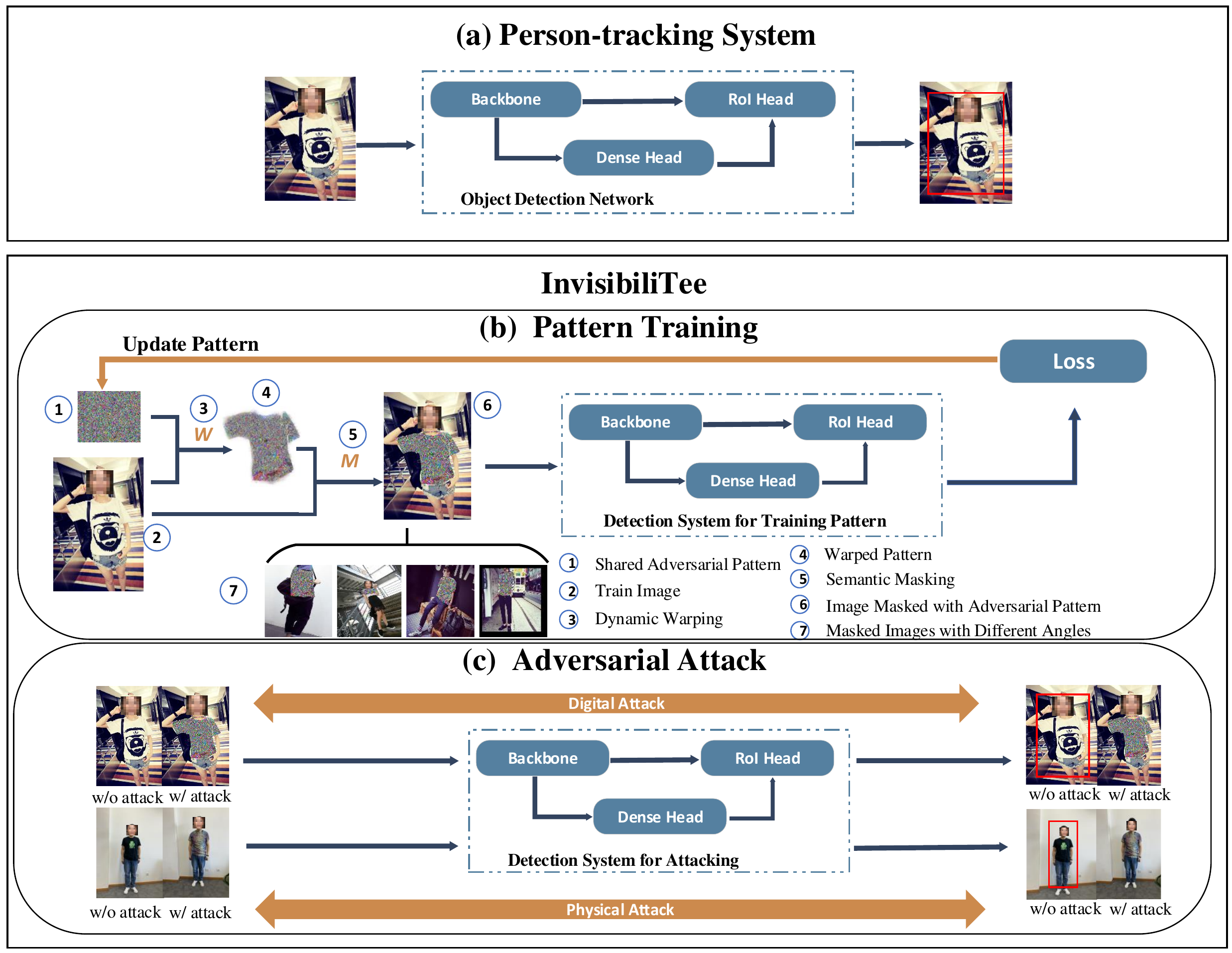} 
    \caption{An overview of normal object detection networks in persons-tracking systems (a) and the framework of InvisibiliTee (b)(c). \textbf{(a)} is a normal detection pipeline. \textbf{(b)} illustrates the process of adversarial pattern training, including dynamic warping, semantic masking and pattern training. \textbf{(c)} describes the process of adversarial attack in both digital and physical scenario.}
    \label{model}
\end{figure*}

\vspace{-0.6cm}
\subsection{Overview}
%This section introduces the \textbf{InvisibiliTee} model, a black-box adversarial attack method on state-of-the-art human detection models.
As illustrated in Fig.~\ref{model}(b),  the InvisbiliTee's network contains two parts: the attack head and object detection system. Attack head is designed to generate the shared adversarial pattern which consists of three steps. 
1, \textbf{Dynamic Warping.} According to the clothing key points annotation, the angle and position of the pattern are adjusted dynamically through warping. 2, \textbf{Semantic Masking.} We obtain the clothing polygon mask from annotation and attach the pattern onto the original image to generate the attack picture. 3, \textbf{Pattern Training.} Adversarial pattern consists of color pixels, which are directly derived from learnable neural network parameters.
During training, only the parameters of adversarial pattern are updated, while the pre-trained detection model remains fixed.
A set of adversarial attacking losses and geometric constraint losses are proposed to update pattern parameters. 

\eat{\vspace{-0.3cm}
\begin{enumerate}
    \item \textbf{Dynamic Warping.}
According to the clothing key points annotation, the angle and position of the pattern are adjusted dynamically through warping.
\vspace{-0.26cm}
\item \textbf{Semantic Masking.}
Obtain the clothing polygon mask from annotation and attach the pattern onto the original image to generate the attack picture.
\vspace{-0.26cm}
\item \textbf{Pattern Training.}
Adversarial pattern consists of color pixels, which are directly derived from learnable neural network parameters.
During training, only the parameters of adversarial pattern are updated, while the pre-trained detection model remains fixed.
A set of adversarial attacking losses and geometric constraint losses are proposed to update pattern parameters. 
\eat{to generate new pattern that can fool the detectors into wrong direction. We also apply Total Variation loss and Non-Printability Score to constrain the pattern. The former makes the adversarial pattern more continuous, whilst the latter makes it printable.}
\end{enumerate}}

%\vspace{-0.3cm}
The second part of the network is object detection system. According to the different processing stages, single-stage framework YOLO~\cite{redmon2016you,redmon2017yolo9000,farhadi2018yolov3} and two-stage framework Faster R-CNN~\cite{ren2016faster} are implemented.
Note that following the practice of previous research~\cite{liu2019dpatch}, the adversarial pattern here is a group of network parameters. So it can be optimized during model training. It is randomly initialized and updated iteratively. The learned adversarial pattern is tested on other unseen models following black-box attack protocols to evaluate the attacking performance, as shown in Fig.~\ref{model}(c). 
\eat{Two types of adversarial attacks are performed: digital and physical attack. In the digital attack, the patterned T-shirt shape is overlaid to replace the clothes in the image, and then attacks unseen models.In the physical attack, the human participants wear real T-shirts with adversarial pattern to attack the object detection systems.On the other hand, as the control group, both the original images in digital experiment and human participant without the adversarial pattern are tested. The results are illustrated in the Fig.~\ref{model} (a).}

\vspace{-0.4cm}
\subsection{Attack and Geometric Constraint Loss Functions}
\label{sec:loss}
\vspace{-0.2cm}
\subsubsection{Attack Loss} 
Attack Losses are optimized to fool the detectors into making wrong decisions.
When training with YOLO architecture, $Loss_{attk}$ consists of three training losses.
\begin{equation}
Loss_{attk}= \theta_{11}Loss_{cla} +  \theta_{12}Loss_{coord} + \theta_{13}Loss_{wh}.
\end{equation}

When training with Faster R-CNN architecture, $Loss_{attk}$ consists of two training losses.
\begin{equation}
Loss_{attk}= \theta_{21}Loss_{cla} +  \theta_{22}Loss_{bbox}.
\end{equation}

Classification cross-entropy attack loss is expressed as
\begin{equation}
Loss_{cla} = \sum_{i=0}^{C-1} y_{i} \log \left(\hat{y}_{i}\right)
\end{equation}
where $y_i$ stands for the ground truth category label, $\hat{y}_{i}$ indicates the probability that current sample belongs to the i-th category and $C$ is the number of categories. 
Bounding box coordinate cross-entropy attack loss is defined as
\begin{equation}
Loss_{coord} = \sum_{i=1}^{2} x_{i} \log \left(\hat{x}_{i}\right)
\end{equation}
where $x_i$ stands for the ground truth bounding box coordinate $x_i = (x_i^1, x_i^2)$, $\hat{x}_{i}$ indicates the predicted one. 
Bounding box width-height mean squared error attack loss is presented as
\begin{equation}
Loss_{wh}=e^{-\frac{1}{2} \left((w-\hat{w})^{2}+(h-\hat{h})^{2}\right)}
\end{equation}
where $w$ and $h$ stand for the width and height of the bounding box respectively.
Bounding box L1 attack loss is presented as
\begin{equation}
Loss_{bbox}=e^{-|\hat{t}_{i}-{t}_{i}|}
\end{equation}
$\hat{t}_{i}$ is a vector representing the 4 parameterized coordinates
of the predicted bounding box, and ${t}_{i}$ is that of the ground-truth box.

\vspace{-0.5cm}
\subsubsection{Geometric Constraint Loss} 
\vspace{-0.2cm}

To learn a more continuous as well as printable adversarial pattern, we introduce total variation loss and non-printability score to constrain the pattern in the training process.

Total Variation (TV) loss encourages spatial smoothness in the generated image. The TV loss is defined as follows,
\begin{equation}
    L_{tv}(P)=\sum_{i, j} \sqrt{\left(\left(a_{i, j}-a_{i+1, j}\right)^{2}+\left(a_{i, j}-a_{i, j+1}\right)^{2}\right.}
\end{equation}

where $P$ is the adversarial pattern, $a_{i,j}$ stands for the pixel value in the pattern of position $(i,j)$. $L_{tv}(P)$ minimizes the distance between neighboring pixels. Thus it can make the pattern more natural and smooth.

%Non-printability Score loss ensures the patterns can be printed.
Non-printability score~\cite{thys2019fooling} reflects how difficult it is to print the pattern, and the lower the value, the less distorted it will be printed. It is defined as:
\begin{equation}
L_{print}(P)=\sum_{a_{(i,j)} \in P} \min _{c_{\text {print }} \in C}\left|a_{(i,j)}-c_{\text {print }}\right|.
\end{equation}
where $c_{\text {print }}$ stands for the pre-defined printable colors.
$L_{print}(P)$ minimizes the distance between pattern pixels and printable pixels. 
\eat{Thus it can make the pattern printable and avoid too bight color.}

The Joint Constraint Loss $Loss_{cons}$ can be represented by the equation below:
\begin{equation}
Loss_{cons}= \alpha_{1} L_{tv}(P) + \alpha_{2} L_{print}(P)
\end{equation}
where $\alpha_{1}$ and $\alpha_{2}$ are hyper-parameters.

\vspace{-0.4cm}
\subsubsection{Overall Loss Function} 
The final loss $L$ of our attack model can be expressed as:
\begin{equation}
L=  Loss_{attk} + Loss_{cons}
\end{equation}

\vspace{-0.5cm}
\subsection{Geometric Warp and Masking}

\begin{figure}[htp]
    \centering
    \includegraphics[width=0.6\textwidth]{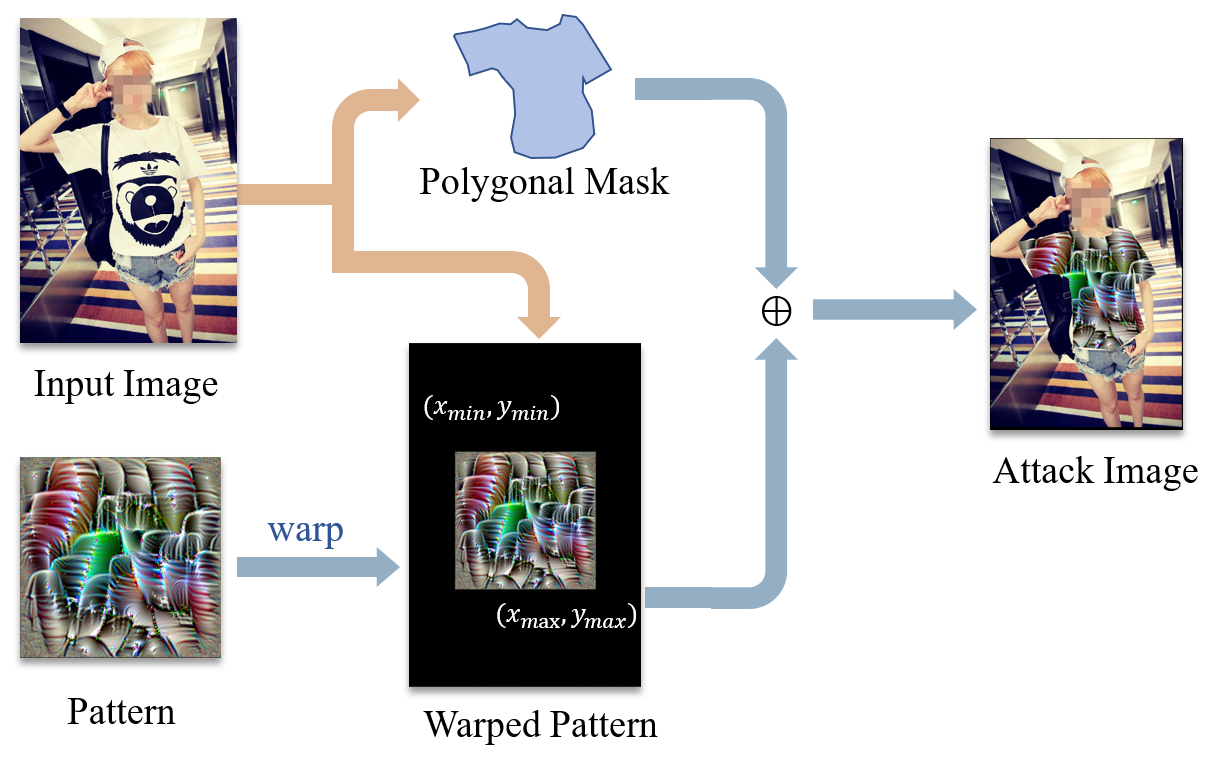} 
    \vspace{-0.4cm}
    \caption{The process of geometric warp and masking.}
    \label{fig:warp}
\end{figure}

\vspace{-0.3cm}
As shown in Fig.~\ref{fig:warp}, we apply geometric warp to the pattern according to the T-shirt key points before attack via a polygonal mask. In Fig.~\ref{fig:warp}, brown lines stand for information supervision while blue lines stand for transformation. 
Geometric warp operation is also known as perspective transformation, widely used to project the image to a new viewing plane, and its general transformation formula is:
\begin{equation}
    \left[\begin{array}{lll}
x' & y' & w'
\end{array}\right]=\left[\begin{array}{lll}
u & v & w
\end{array}\right]  \cdot T
\end{equation}

$(u, v)$ are the original image pixel coordinates, $(x=x'/w', y=y'/w')$ are the transformed image pixel coordinates. $T$ is the perspective transformation matrix.
Given corresponding four pairs of pixel coordinates before/after perspective transformation, the perspective transformation matrix $T$ can be obtained.
We warp the pattern image to fit the input image, so the original 4 paires of coordinates are the 4 corners of the pattern. 
%More details are shown in supplementary material.

After geometric warping, we use the polygonal mask $M$ formed by the T-shirt key points to attach the pattern onto the image, replacing the original T-shirt outward appearance.
\begin{equation}
    I_{attk} = (1 - Mask)\odot I_{ori} + M \odot P
\end{equation}

\vspace{-0.3cm}
\section{Attacks in the Digital World} \label{sec:exp-digital}
\vspace{-0.2cm}

\subsection{Dataset and Experiment Setup}
The Deep Fashion2 dataset~\cite{DeepFashion2} is a comprehensive fashion dataset containing pictures of people wearing various types of clothing with annotations of clothing key points. Each image has around 15-30 key points to outline the clothes, which is shown in Fig.6. We leverage the open sourced object detection project MMDetection~\cite{mmdetection} and pre-trained models to implement our experiments.
YOLOv3~\cite{farhadi2018yolov3} and Faster R-CNN~\cite{ren2016faster} are chosen as the training detection model. 
%Please refer to supplementary material for detail data set preprocessing and hyper-parameters settings.
The adversarial pattern is trained on the modified Deep Fashion2 dataset with Adam optimizer with initial learning rate $1e^{-3}$. The hyper-parameters $\theta_{11} \sim \theta_{13}$, $\alpha_1$ and $\alpha_2$ are $5$, $1$, $1$, $100$, $100$ respectively in YOLOv3. And $\theta_{21}, \theta_{22}$, $\alpha_1$ and $\alpha_2$ are $500$, $10$, $18$, $100$ in Faster R-CNN, balancing the loss terms into the same scale. 
Fig.5 shows the changing of total loss during training.

\begin{table}[h]
\small
\centering
\caption{Dataset information after reconstruction. }
%\vspace{-0.2cm}
\label{tab:dataset}
\begin{tabular}{l|r|r|r|r}
\toprule[1pt]
\textbf{Split}  & \multicolumn{1}{l|}{\textbf{No human}} & \multicolumn{1}{l|}{\begin{tabular}[c]{@{}l@{}}\textbf{Frontal Viewpoint} \end{tabular}} & \multicolumn{1}{l|}{\begin{tabular}[c]{@{}l@{}} \textbf{Size/back Viewpoint} \end{tabular}} & \multicolumn{1}{l}{\textbf{Total}} \\ \hline
\textbf{Train}    & 6,664                         & 56,121                                                                              & 8,860                                                                                 & 64,981                     \\ 
%\textbf{Included} & ×                             & \checkmark                                   &\checkmark                                                                                     &                            \\ \hline
\textbf{Val}      & 1,027                         & 9,855                                                                               & 1,674                                                                                 & 11,529                     \\ 
%\textbf{Included} & ×                             & \checkmark                                                                                 & \checkmark                                                                                     &                            \\ \hline
\bottomrule[1pt]
\end{tabular}
\end{table}

\begin{figure*}[t]
    \label{fig:loss}
    \centering
    \begin{minipage}[t]{0.6\textwidth}
    \centering
    \includegraphics[width=0.9\textwidth]{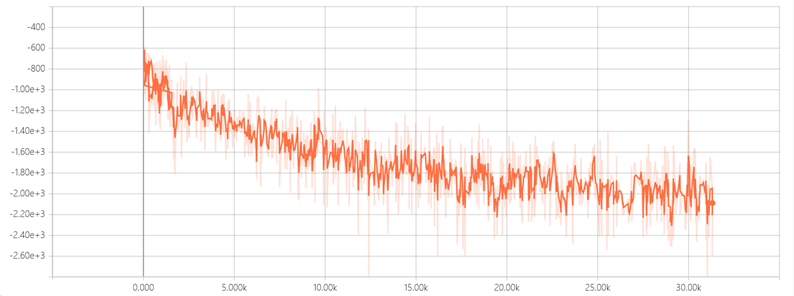} 
    \caption{Train loss with YOLO architecture.}
    \end{minipage}
    \begin{minipage}[t]{0.3\textwidth}
    \centering
    \includegraphics[width=0.6\textwidth]{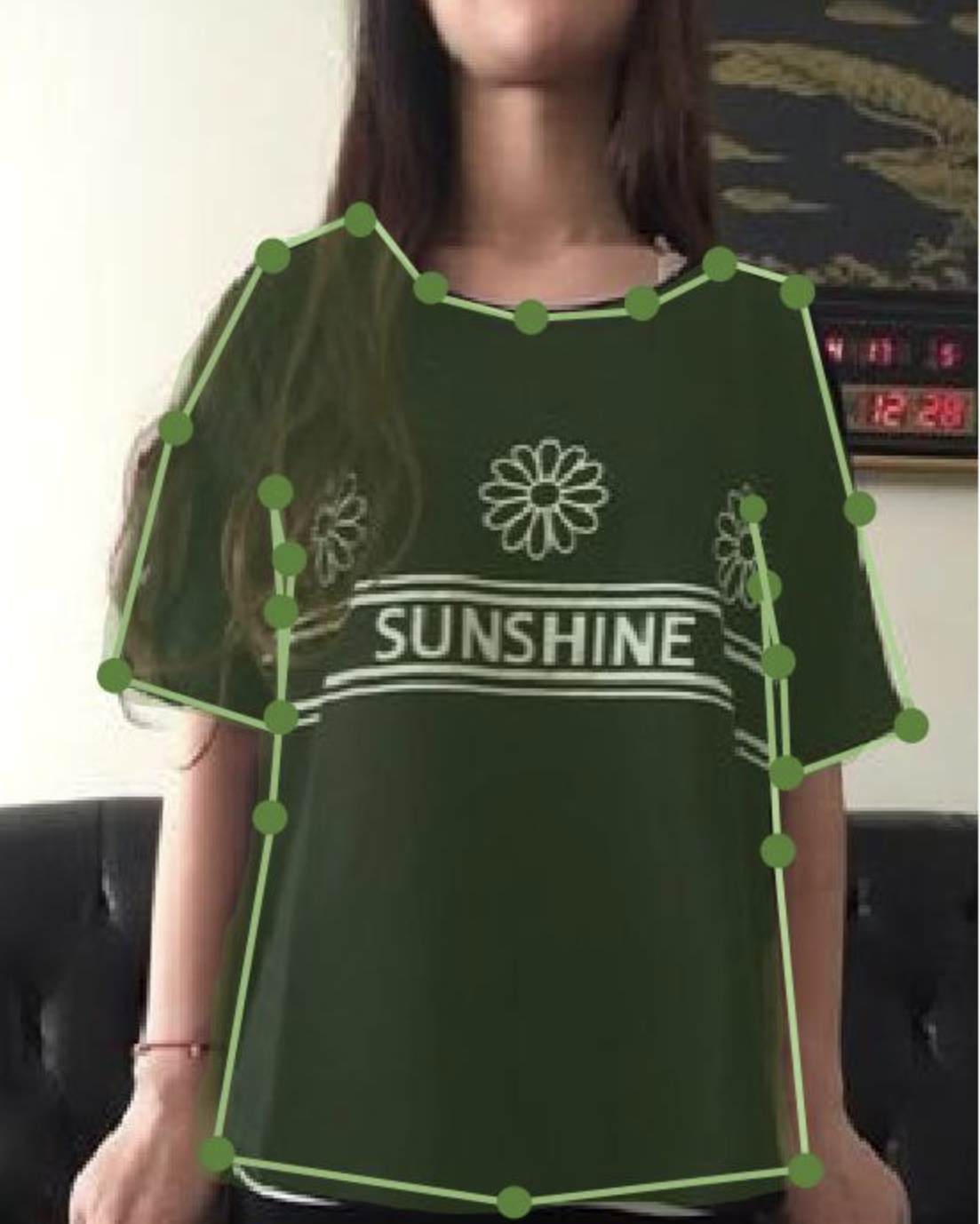}
    \caption{Key point annotations in one image}
    \end{minipage}
\end{figure*}

\vspace{-0.4cm}
\subsection{Experimental Results}
\vspace{-0.2cm}

We train on a specific object detection model and then transfer the trained pattern to attack other unseen models.
%It is considered as a black-box attack since the knowledge of the target detector parameters are unknown during the training phase.

\subsubsection{Black-box Attacks.} 
\vspace{-0.2cm}

\vspace{-0.6cm}
\begin{table*}[ht]
\small
\centering
\caption{The black-box (digital) attack results on SOTA detectors with an adversarial pattern trained on YOLOv3~\cite{farhadi2018yolov3} and Faster-RCNN~\cite{ren2016faster}. 
The patterns used are shown in Fig.~\ref{fig:question} A and B respectively. 
}
% \vspace{-0.2cm}
% \hspace{-0.4cm}
\resizebox{\textwidth}{!}
{
\begin{tabular}{@{}c|ccc|ccc|ccc@{}}
\toprule[1pt]
\multirow{2}{*}{\textbf{\begin{tabular}[c]{@{}c@{}} \\  \\ Target model\end{tabular}}} & \multicolumn{3}{c|}{\textbf{AP@IoU=0.50:0.95}}                                                                                                                                    & \multicolumn{3}{c|}{\textbf{AP@0.50}}                                                                                                                                             & \multicolumn{3}{c}{\textbf{AP@0.75}}                                                                                                                                              \\ \cmidrule(l){2-10} 
                                       & \textbf{\begin{tabular}[c]{@{}c@{}}without \\ attack\end{tabular}} & \textbf{\begin{tabular}[c]{@{}c@{}}Faster R-CNN\\ pattern attack\end{tabular}} & \textbf{\begin{tabular}[c]{@{}c@{}}YOLO\\ pattern\\ attack\end{tabular}} & \textbf{\begin{tabular}[c]{@{}c@{}}without \\ attack\end{tabular}} & \textbf{\begin{tabular}[c]{@{}c@{}}Faster R-CNN\\ pattern attack\end{tabular}} & \textbf{\begin{tabular}[c]{@{}c@{}}YOLO\\ pattern\\ attack\end{tabular}} & \textbf{\begin{tabular}[c]{@{}c@{}}without \\ attack\end{tabular}} & \textbf{\begin{tabular}[c]{@{}c@{}}Faster R-CNN\\ pattern attack\end{tabular}} & \textbf{\begin{tabular}[c]{@{}c@{}}YOLO\\ pattern\\ attack\end{tabular}} \\ \midrule
YOLOv3~\cite{farhadi2018yolov3}                                 & 0.254                   & 0.204                                                                          & 0.003                                                                  & 0.601                   & 0.512                                                                          & 0.022                                                                  & 0.178                   & 0.131                                                                          & 0.000                                                                  \\ 
SSD~\cite{liu2016ssd}                                  & 0.256                   &0.184                                                                                &0.187                                                                        & 0.598                   &0.532                                                                                &0.558                                                                        & 0.133                   &0.100                                                                                &0.102                                                                        \\
RetinaNet~\cite{lin2017focal}               & 0.216                   & 0.000                                                                          &0.000                                                                        & 0.607                   & 0.000                                                                          &0.000                                                                        & 0.120                   & 0.000                                                                          &0.000                                                                        \\ 
\cline{1-10}
Faster-RCNN~\cite{ren2016faster}                         & 0.213                   & 0.001                                                                          & 0.177                                                                  & 0.564                   & 0.007                                                                          & 0.512                                                                  & 0.128                   & 0.000                                                                          & 0.090                                                                  \\
Faster-RCNN+softnms                    & 0.217                   & 0.041                                                                          &0.087                                                                        & 0.573                   & 0.164                                                                          &0.270                                                                        & 0.131                   & 0.009                                                                          &0.042                                                                        \\
Mask R-CNN ~\cite{matterport_maskrcnn_2017}                       & 0.210                   & 0.047                                                                          &0.071                                                                        & 0.573                   & 0.197                                                                          &0.256                                                                       & 0.117                   & 0.010                                                                          &0.026                                                                        \\
Cascade Mask R-CNN~\cite{Cai_2019}                   & 0.221                   & 0.063                                                                          &0.089                                                                        & 0.599                   & 0.232                                                                          &0.291                                                                        & 0.124                   & 0.019                                                                          &0.041                                                                        \\
Cascade R-CNN~\cite{Cai_2019}                        & 0.23                    & 0.038                                                                          & 0.061                                                                  & 0.607                   & 0.159                                                                          & 0.216                                                                  & 0.132                   & 0.010                                                                          & 0.031                                                                  \\
Dynamic R-CNN~\cite{DynamicRCNN}                     & 0.227                   & 0.053                                                                          & 0.078                                                                  & 0.600                   & 0.202                                                                          & 0.264                                                                  & 0.133                   & 0.015                                                                          & 0.035                                                                  \\
\cline{1-10}
CornerNet~\cite{law2018cornernet}                             & 0.222                   & 0.153                                                                          & 0.084                                                                  & 0.541                   & 0.438                                                                          & 0.249                                                                  & 0.157                   & 0.084                                                                          & 0.044                                                                  \\
RepPoints~\cite{yang2019reppoints}                             & 0.369                   & 0.046                                                                          & 0.067                                                                  & 0.567                   & 0.190                                                                          & 0.252                                                                  & 0.397                   & 0.011                                                                          & 0.023                                                                  \\
FreeAnchor~\cite{zhang2019freeanchor}                          & 0.218                   & 0.000                                                                          & 0.078                                                                  & 0.598                   & 0.000                                                                          & 0.265                                                                  & 0.121                   & 0.000                                                                          & 0.038                                                                  \\
SABL~\cite{Wang_2020_ECCV}                                & 0.265                   & 0.058                                                                          & 0.076                                                                  & 0.657                   & 0.210                                                                          & 0.249                                                                  & 0.175                   & 0.020                                                                          & 0.040                                                                  \\
PAA~\cite{paa-eccv2020}                                 & 0.243                   & 0.075                                                                          & 0.089                                                                  & 0.607                   & 0.237                                                                          & 0.276                                                                  & 0.178                   & 0.036                                                                          & 0.045                                                                  \\

\bottomrule[1pt]
\end{tabular}
}

\label{tab:sota}
\end{table*}

Table~\ref{tab:sota} shows the black-box attack performance on several different SOTA detection models, where AP stands for Average Precision, the most commonly used evaluation metric for target detection.  Our attack model is able to significantly reduce AP at different IoU ratios.
%The first column in Table 2 shows the name of the assumed target pedestrian detection model.
\eat{We tested on 14 models in total to verify whether the adversarial pattern can mislead these detectors and make it unable to accurately detect pedestrians.
The changes of $AP@IoU$ before and after the attack ia coompared, which is the most commonly used evaluation metric for target detection.}
The $AP@IoU=0.50:0.95$ corresponds to average mAP calculated over a range of IoU thresholds, from 0.50 to 0.95 with the uniform step size 0.05.
%, which is used as the primary metric to evaluate the object detectors.
As shown in column 2 and 3, the $AP@IoU=0.50:0.95$ is reduced from about $0.25$ to about $0.08$. 
%These shows that the adversarial pattern greatly reduces the probability of people being detected. 
%We have also shown results for $AP@IoU=0.5$ and $AP@IoU=0.75$ for detail comparison.
In column 5, it can be seen that even under the most relaxed threshold condition ($IoU > 0.5$ means that it was a hit ), our method reduces the evaluation to less than half in most models.
When the IoU threshold is set to $0.75$, it can be considered that the detector is close to completely invalid after attack.
%The experiment of black-box attack shows that our method can attack the target detect models successfully without knowing the information about them.

% \vspace{-0.9cm}
\subsubsection{Effect of Different Pattern Resolution} 
\vspace{-0.2cm}

\begin{table*}[htpb]
\centering
\small
\caption{Black-box experiment results of different pattern resolutions trained on Faster-RCNN. ``w/o attack'' means no attack is performed. P 200, p 100, p 50 stand for adversarial pattern resolution of $200\times200$, $100\times100$ and $50\times50$ pixels, respectively.}

\resizebox{\textwidth}{!}
{
\begin{tabular}{c|c|ccccccc}
\toprule[1pt]
\multicolumn{2}{c|}{\textbf{target   model}}                                 & \textbf{CornerNet} & \textbf{\begin{tabular}[c]{@{}c@{}}Cascade\\       R-CNN\end{tabular}} & \textbf{\begin{tabular}[c]{@{}c@{}}Mask \\      R-CNN\end{tabular}} & \textbf{\begin{tabular}[c]{@{}c@{}}Faster\\      R-CNN\end{tabular}} & \textbf{\begin{tabular}[c]{@{}c@{}}Faster-RCNN\\      +softnms\end{tabular}} & \textbf{\begin{tabular}[c]{@{}c@{}}Cascade \\      Mask R-CNN\end{tabular}} & \textbf{YOLOv3} \\ \hline
\multicolumn{1}{c|}{\multirow{4}{*}{\textbf{AP@0.50}}} & \textbf{w/o attack} & 0.541              & 0.607                                                                  & 0.573                                                               & 0.564                                                                & 0.573                                                                        & 0.599                                                                       & 0.601           \\ \cline{2-9} 
\multicolumn{1}{c|}{}                                  & \textbf{p 200}      & 0.438              & 0.159                                                                  & 0.197                                                               & 0.007                                                                & 0.164                                                                        & 0.232                                                                       & 0.512           \\
\multicolumn{1}{c|}{}                                  & \textbf{p  100}     & 0.437              & 0.150                                                                  & 0.165                                                               & 0.126                                                                & 0.136                                                                        & 0.208                                                                       & 0.529           \\
\multicolumn{1}{c|}{}                                  & \textbf{p  50}      & 0.443              & 0.185                                                                  & 0.211                                                               & 0.174                                                                & 0.188                                                                        & 0.289                                                                       & 0.524           \\
\bottomrule[1pt]                                                                        
\end{tabular}
}
\label{tab:patt_size}
\end{table*}

\begin{figure}[h]
    %\vspace{-0.4cm}
    \centering
    \includegraphics[width=0.95\textwidth]{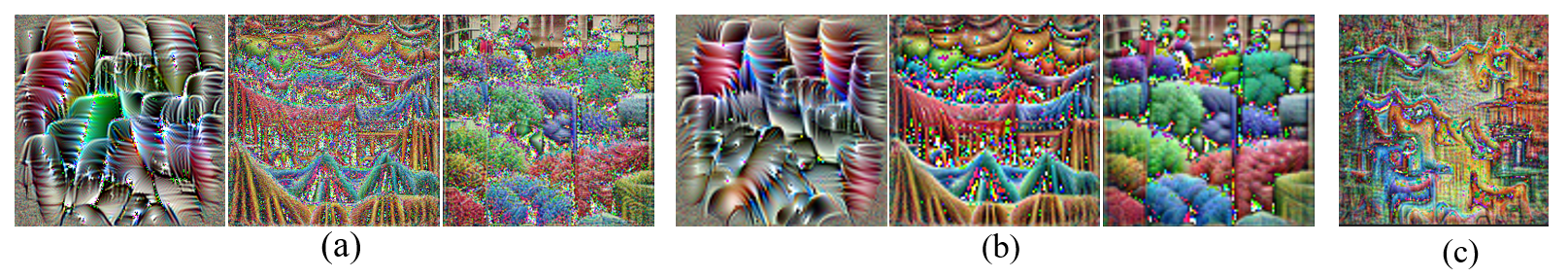} 
    \vspace{-0.3cm}
    \caption{Adversarial patterns trained with different resolution in pixels. (a) and (b) are random initialized while (c) is initialized by a texture.}
    \label{fig:pattern}
\end{figure}

\vspace{-0.4cm}
\begin{figure}[h]
    %\vspace{-0.4cm}
    \centering
    \includegraphics[width=0.8\textwidth]{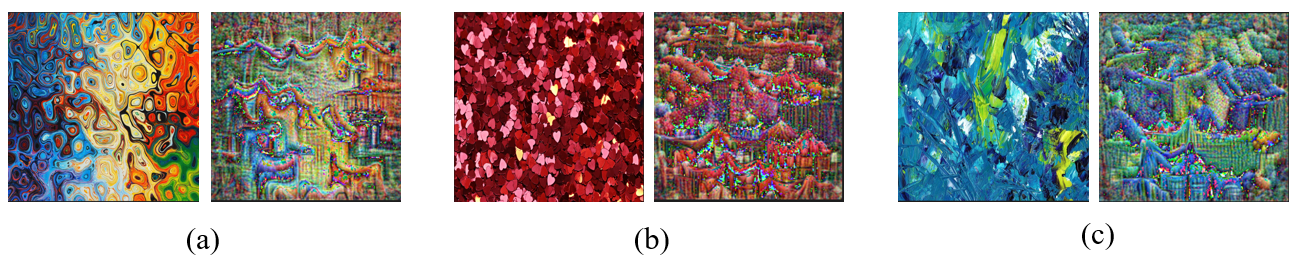} 
    \vspace{-0.3cm}
    \caption{Adversarial patterns initialized by different textures. In each group, the left one is the texture and the right one is the corresponding adversarial pattern.}
    \label{fig:texturep}
\end{figure}
%\vspace{-0.2cm}

Fig. ~\ref{fig:pattern} shows several adversarial patterns with different resolutions. (a) are patterns in size $200\times200$ while (b) are corresponding patterns in size $100\times100$. The first pattern on left is trained on Fsater-RCNN and the last two are trained on YOLOv3. Fig.~\ref{fig:texturep} shows everal adversarial patterns initialized by different textures.
 We use downsampling in training to study the effect of image resolution. These corresponding patterns with different resolutions are firstly random initialized in size $400\times400$ with the same random seed and then they are downsampled to different resolutions before dynamic warping.
As Table~\ref{tab:patt_size} indicates, high-resolution adversarial pattern performs slightly better in most cases, but the advantage is not obvious. It is believed that a relatively lower resolution can be better considering a limited scale of training set. 
%More results can be seen in supplementary material. 

\vspace{-0.5cm}
\subsection{Case Studies of Digital Attacks}
\vspace{-0.3cm}

\begin{figure}[htbp]
    \centering
    \includegraphics[width=0.95\textwidth]{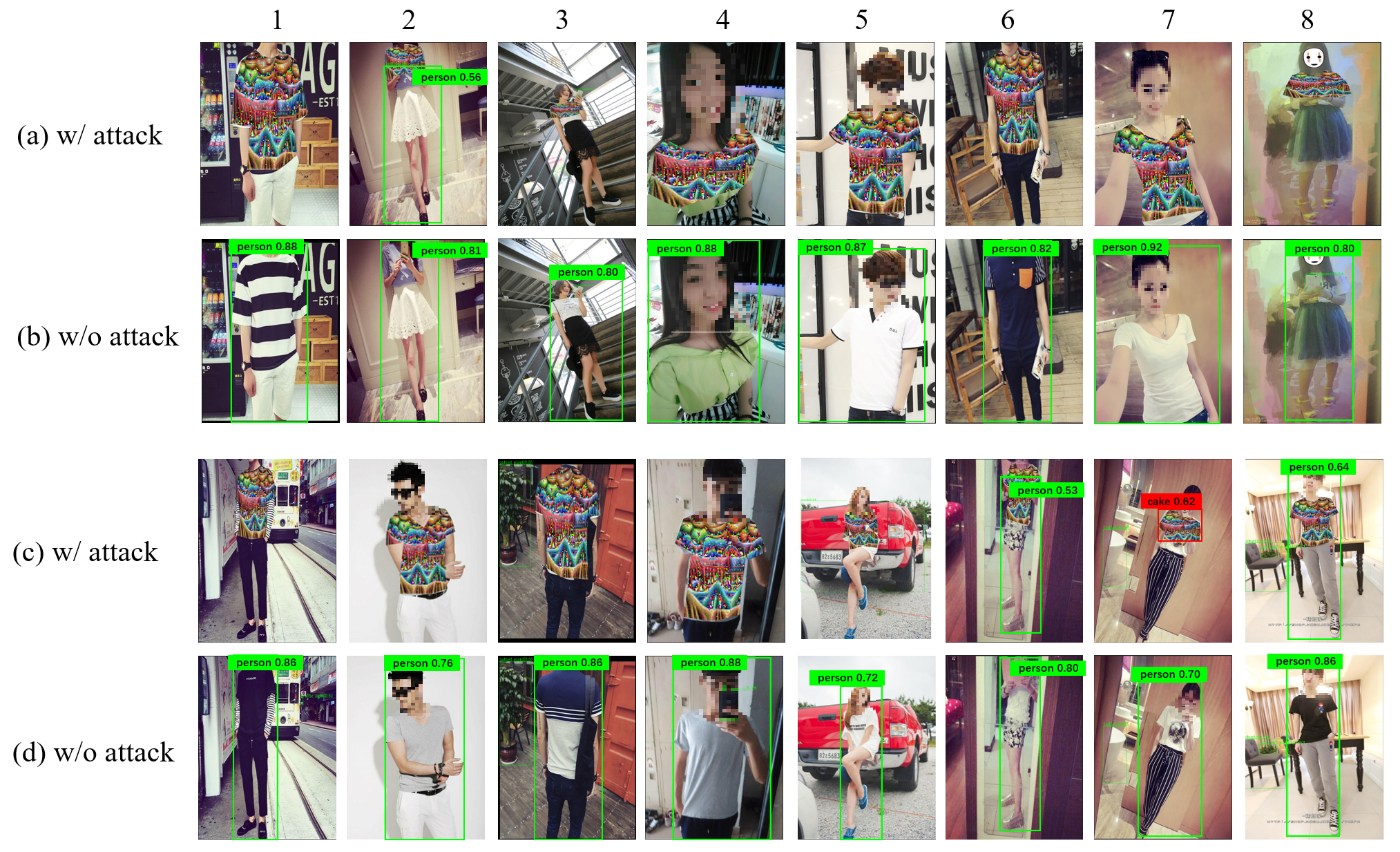} 
    
    \caption{Digital attack results. ``w/ attack'' and ``w/o attack'' stand for with attack and without attack respectively. }
    \label{fig:digital_eg}
\end{figure}

Fig.~\ref{fig:digital_eg} shows the digital attack results when we use the adversarial pattern trained on YOLOv3~\cite{farhadi2018yolov3} to attack the target model CornerNet~\cite{law2018cornernet}. Only bounding boxes with a confidence score over 0.5 are shown.
As Fig.~\ref{fig:digital_eg} indicates, successful attack results can be roughly divided into following three types, which are namely detection failure, bounding box miss and category error. Most cases belong to the first type, which are (a)1, (a)3-8 and (c)1-5. (a)2 and (c)6 provide a miss-leading or incomplete bounding box. In (c)7, the target model recognizes it as a cake instead of a person. There are also some failure cases as (c)8. However, wearing a ``InvisibiliTee'' still manages to make the confidence score drop from 0.86 to 0.64.

\vspace{-0.5cm}
\section{Attacks in the Physical World} \label{sec:exp-physical}

\vspace{-0.2cm}
We print the adversarial pattern on fabric and tailor it into an ``InvisibiliTee'' to evaluate attack effects in physical world.

%\vspace{-0.4cm}
%\subsection{Physical Experiment}
%\vspace{-0.2cm}

% \vspace{-0.3cm}
\begin{figure}[htbp]
    \centering
    \includegraphics[width=0.95\textwidth]{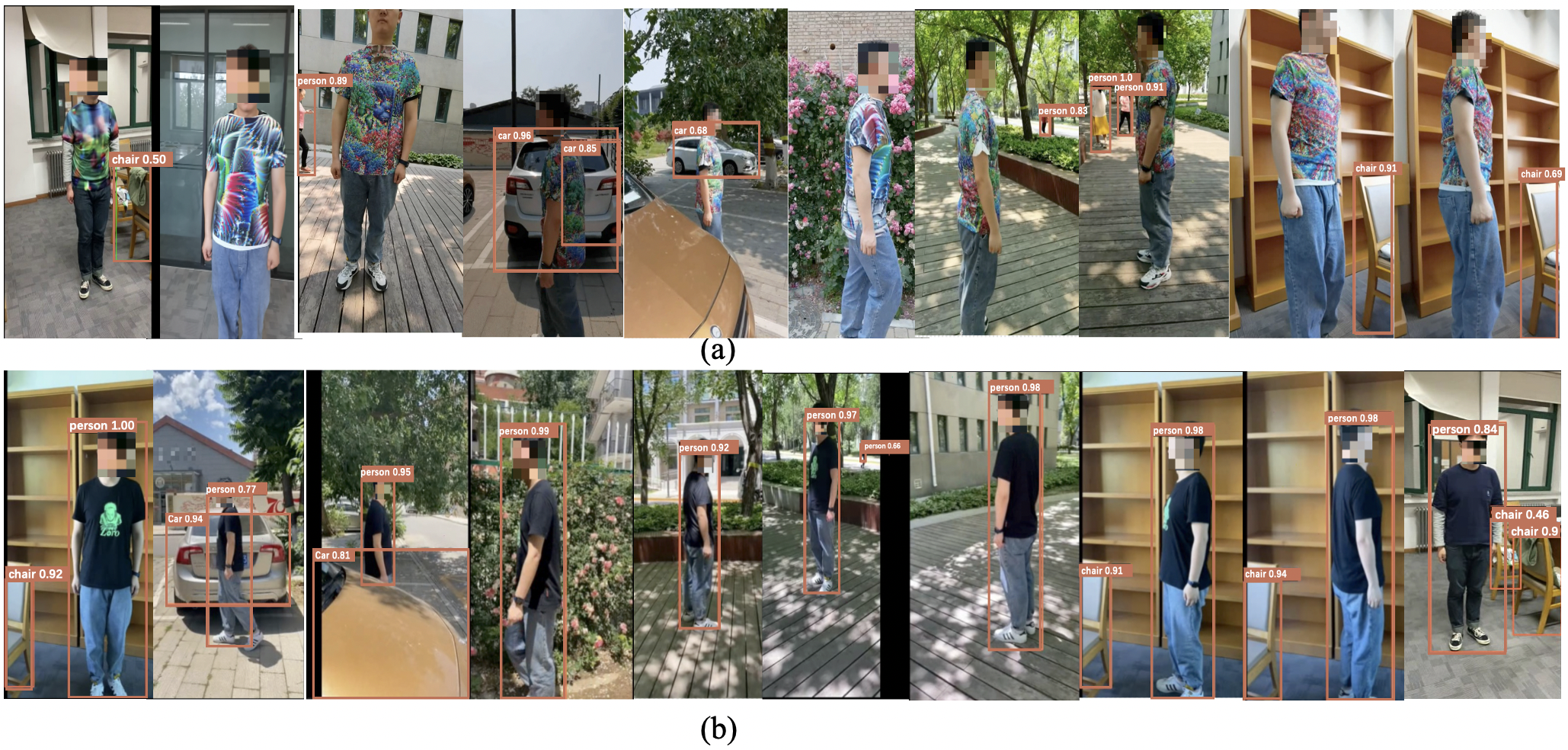} 
    \vspace{-0.4cm}
    \caption{Adversarial attack in real world. In (a), the pedestrian wears an ``InvisibiliTee'' while (b) not. }
    \label{physical attack}
\end{figure}

Fig.~\ref{physical attack} shows a typical scene in real life when a pedestrian walked from a outdoor parking lot into a building. Note that the mosaics are added for anonymity only and do not have any impact in the experiments.
When he wears an ``InvisibiliTee'', detectors fail to identify the person. But at the same time it is able to recognize other objects or even people. 
%However, the pedestrian can be identified easily without an ``InvisibiliTee'' on.  

\vspace{-0.5cm}
\subsection{Additional Discussion}
\vspace{-0.2cm}
There are perceived security concerns for developing technologies to cloak from person-tracking systems, for potential malicious uses. Although this is a valid concern to some degree, we argue that the risk is manageable for the following reasons:
1, The adversarial patterns are generated in a way that makes them easily differentiable from daily clothing, almost making a public statement that the wearer does not intend to be detected by person-tracking systems. They can be extremely noticeable and therefore discouraging for malicious parties to wear.
2, In critical areas where additional security is required, the InvisibiliTee model could be used in adversarial training to improve robustness against such attacks.

\vspace{-0.4cm}
\section{Conclusion} \label{sec:conclusion}
\vspace{-0.2cm}

% What
This paper presented a black-box attack method, \textbf{InvisibiliTee}, to perform digital and physical adversarial attacks on the human detection models for individual privacy preservation.
% How
An angle-agnostic scheme trained by the attack and geometric constraint losses was proposed to generate adversarial patterns. 
%The learned patterns were warped and overlaid on the T-shirt appeared in the images to simulate a person that wearing an ``InvisibiliTee'' in real world.
% Result, contribution
Both digital and physical attacking experiments were conducted on a group of state-of-the-art human detection systems, demonstrating the effectiveness of the learned adversarial patterns. The results have shown that the \textbf{InvisibiliTee} can significantly reduce the average precision of person detection systems especially in the digital attacks. In addition, 3D adversarial attack is an emerging research. Therefore, it is challenging to extent \textbf{InvisibiliTee} to a 3D model.

\vspace{-0.4cm}
\bibliographystyle{splncs04}
\bibliography{samplepaper}

\begin{thebibliography}{10}
\providecommand{\url}[1]{\texttt{#1}}
\providecommand{\urlprefix}{URL }
\providecommand{\doi}[1]{https://doi.org/#1}

\bibitem{brown2017adversarial}
Brown, B.T., Mané, D., Roy, A., Abadi, M., Gilmer, J.: Adversarial patch. CoRR
   (2017)

\bibitem{Cai_2019}
Cai, Z., Vasconcelos, N.: Cascade r-cnn: High quality object detection and
  instance segmentation. TPAMI p. 1–1 (2019).
  \doi{10.1109/tpami.2019.2956516},
  \url{http://dx.doi.org/10.1109/tpami.2019.2956516}

\bibitem{mmdetection}
Chen, K., Wang, J., Pang, J., Cao, Y., Xiong, Y., Li, X., Sun, S., Feng, W.,
  Liu, Z., Xu, J., Zhang, Z., Cheng, D., Zhu, C., Cheng, T., Zhao, Q., Li, B.,
  Lu, X., Zhu, R., Wu, Y., Dai, J., Wang, J., Shi, J., Ouyang, W., Loy, C.C.,
  Lin, D.: {MMDetection}: Open mmlab detection toolbox and benchmark. arXiv
  preprint arXiv:1906.07155  (2019)

\bibitem{dai2016r}
Dai, J., Li, Y., He, K., Sun, J.: R-fcn: object detection via region-based
  fully convolutional networks. In: NeurIPS. pp. 379--387 (2016)

\bibitem{farhadi2018yolov3}
Farhadi, A., Redmon, J.: Yolov3: An incremental improvement. In: CVPR. pp.
  1804--2767. Springer Berlin/Heidelberg, Germany (2018)

\bibitem{fu2017dssd}
Fu, C.Y., Liu, W., Ranga, A., Tyagi, A., Berg, C.A.: Dssd : Deconvolutional
  single shot detector. CVPR  (2017)

\bibitem{DeepFashion2}
Ge, Y., Zhang, R., Wu, L., Wang, X., Tang, X., Luo, P.: A versatile benchmark
  for detection, pose estimation, segmentation and re-identification of
  clothing images. CVPR  (2019)

\bibitem{girshick2015fast}
Girshick, R.: Fast r-cnn. In: CVPR. pp. 1440--1448 (2015)

\bibitem{goodfellow2014explaining}
Goodfellow, J.I., Shlens, J., Szegedy, C.: Explaining and harnessing
  adversarial examples. ICLR  (2014)

\bibitem{matterport_maskrcnn_2017}
He, K., Gkioxari, G., Doll{\'a}r, P., Girshick, R.: Mask r-cnn. In: CVPR. pp.
  2961--2969 (2017)

\bibitem{paa-eccv2020}
Kim, K., Lee, H.S.: Probabilistic anchor assignment with iou prediction for
  object detection. In: ECCV (2020)

\bibitem{law2018cornernet}
Law, H., Deng, J.: Cornernet: Detecting objects as paired keypoints. In: ECCV
  2018. pp. 765--781. Springer Verlag (2018)

\bibitem{lee2019on}
Lee, M., Kolter, Z.: On physical adversarial patches for object detection. CoRR
   (2019)

\bibitem{lin2017focal}
Lin, T.Y., Goyal, P., Girshick, B.R., He, K., Dollár, P.: Focal loss for dense
  object detection. ICCV pp. 318--327 (2017)

\bibitem{liu2016ssd}
liu, w., anguelov, d., erhan, d., szegedy, c., reed, s.: Ssd: Single shot
  multibox detector. ECCV pp. 21--37 (2016)

\bibitem{liu2019dpatch}
Liu, X., Yang, H., Liu, Z., Song, L., Chen, Y., Li, H.: Dpatch - an adversarial
  patch attack on object detectors. SafeAI@AAAI  (2019)

\bibitem{madry2018towards}
Madry, A., Makelov, A., Schmidt, L., Tsipras, D., Vladu, A.: Towards deep
  learning models resistant to adversarial attacks. ICLR  (2018)

\bibitem{redmon2016you}
redmon, j., Divvala, K.S., Girshick, B.R., farhadi, a.: You only look once:
  Unified, real-time object detection. CVPR  (2016)

\bibitem{redmon2017yolo9000}
Redmon, J., Farhadi, A.: Yolo9000: Better, faster, stronger. CVPR  (2017)

\bibitem{ren2016faster}
Ren, S., He, K., Girshick, R., Sun, J.: Faster r-cnn: towards real-time object
  detection with region proposal networks. TPAMI  \textbf{39}(6),  1137--1149
  (2016)

\bibitem{szegedy2014intriguing}
Szegedy, C., Zaremba, W., Sutskever, I., Bruna, J., Erhan, D., Goodfellow,
  J.I., Fergus, R.: Intriguing properties of neural networks. ICLR  (2014)

\bibitem{thys2019fooling}
Thys, S., Van~Ranst, W., Goedem{\'e}, T.: Fooling automated surveillance
  cameras: adversarial patches to attack person detection. In: CVPR (2019)

\bibitem{uijlings2013selective}
Uijlings, J.R., Van De~Sande, K.E., Gevers, T., Smeulders, A.W.: Selective
  search for object recognition. IJCV  \textbf{104}(2),  154--171 (2013)

\bibitem{Wang2020CVPR}
Wang, H., Wang, G., Li, Y., Zhang, D., Lin, L.: Transferable, controllable, and
  inconspicuous adversarial attacks on person re-identification with deep
  mis-ranking. In: CVPR (June 2020)

\bibitem{Wang_2020_ECCV}
Wang, J., Zhang, W., Cao, Y., Chen, K., Pang, J., Gong, T., Shi, J., Loy, C.C.,
  Lin, D.: Side-aware boundary localization for more precise object detection.
  In: ECCV (2020)

\bibitem{wu2020making}
Wu, Z., Lim, S.N., Davis, L.S., Goldstein, T.: Making an invisibility cloak:
  Real world adversarial attacks on object detectors. In: ECCV. pp. 1--17.
  Springer (2020)

\bibitem{yang2019reppoints}
Yang, Z., Liu, S., Hu, H., Wang, L., Lin, S.: Reppoints: Point set
  representation for object detection. In: ICCV (Oct 2019)

\bibitem{DynamicRCNN}
Zhang, H., Chang, H., Ma, B., Wang, N., Chen, X.: Dynamic r-cnn: Towards high
  quality object detection via dynamic training. In: ECCV. pp. 260--275.
  Springer (2020)

\bibitem{zhang2019freeanchor}
Zhang, X., Wan, F., Liu, C., Ji, R., Ye, Q.: {FreeAnchor}: Learning to match
  anchors for visual object detection. In: NeurIPS (2019)

\bibitem{Zheng2020Reid}
Zheng, Y., Lu, Y., Velipasalar, S.: An effective adversarial attack on person
  re-identification in video surveillance via dispersion reduction. IEEE Access
   \textbf{8},  183891--183902 (2020)

\end{thebibliography}

%Author, F.: Article title. Journal \textbf{2}(5), 99--110 (2016)

%\end{thebibliography}

\end{document}